\documentclass[sigconf]{acmart}

\usepackage{pifont}
\usepackage{multirow}
\usepackage{makecell}
\AtBeginDocument{%
  \providecommand\BibTeX{{%
    \normalfont B\kern-0.5em{\scshape i\kern-0.25em b}\kern-0.8em\TeX}}}

\setcopyright{acmcopyright}
\copyrightyear{2025}
\acmYear{2025}
\acmDOI{XXXXXXX.XXXXXXX}

\acmConference [MM '25]{Proceedings of the 33th ACM International Conference on Multimedia}{October 27-October 31, 2025}{Dublin, Ireland}
\acmBooktitle{Proceedings of the 33th ACM International Conference on Multimedia (MM '25)}
\acmPrice{15.00}
\acmISBN{978-1-4503-XXXX-X/18/06}

\begin{document}


\title{IntentVCNet: Bridging Spatio-Temporal Gaps for Intention-Oriented Controllable Video Captioning}

\author{Tianheng Qiu\textsuperscript{*}}
\affiliation{%
  \institution{University of Science and Technology of China}
  \city{Hefei}
  \country{China}}
\affiliation{%
  \institution{Hefei Institutes of Physical Science,Chinese Academy of Sciences}
  \city{Hefei}
  \country{China}}
\email{thqiu.cs@mail.ustc.edu.cn}

\author{Jingchun Gao\textsuperscript{*}}
\affiliation{%
  \institution{University of Science and Technology of China}
  \city{Hefei}
  \country{China}}
\email{gaojc0714@mail.ustc.edu.cn}

\author{Jingyu Li\textsuperscript{\textdagger}}
\affiliation{%
  \institution{Institute of Artificial Intelligence, Hefei Comprehensive National Science Center}
  \city{Hefei}
  \country{China}}
\affiliation{%
  \institution{State Key Lab. for Novel Software Technology, Nanjing University}
  \city{Nanjing}
  \country{China}}
\email{jingyuli@iai.ustc.edu.cn}

\author{Huiyi Leong}
\affiliation{%
  \institution{University of Chicago}
  \city{Chicago}
  \country{America}}
\email{Joyce.yong@uchicago.edu}  

\author{Xuan Huang}
\affiliation{%
  \institution{Hefei Institutes of Physical Science,Chinese Academy of Sciences}
  \city{Hefei}
  \country{China}}
\email{huangxuan@iim.ac.cn}

\author{Xi Wang}
\affiliation{%
  \institution{National University of Defense Technology}
  \city{Changsha}
  \country{China}}
\email{wx_23ndt@nudt.edu.cn}

\author{Xiaocheng Zhang}
\affiliation{%
  \institution{Harbin Institute of Technology}
  \city{Harbin}
  \country{China}}
\email{22s136029@stu.hit.edu.cn}

\author{Kele Xu}
\affiliation{%
  \institution{National University of Defense Technology}
  \city{Changsha}
  \country{China}}
\email{kele.xu@ieee.org}

\author{Lan Zhang}
\affiliation{%
  \institution{University of Science and Technology
of China}
  \city{Hefei}
  \country{China}}
\email{zhanglan@ustc.edu.cn}

\thanks{*Equal contribution. \textdagger Corresponding author.}





\begin{abstract}
Intent-oriented controlled video captioning aims to generate targeted descriptions for specific targets in a video based on customized user intent.
Current Large Visual Language Models (LVLMs) have gained strong instruction following and visual comprehension capabilities.
Although the LVLMs demonstrated proficiency in spatial and temporal understanding respectively, it was not able to perform fine-grained spatial control in time sequences in direct response to instructions. This substantial spatio-temporal gap complicates efforts to achieve fine-grained intention-oriented control in video.
Towards this end, we propose a novel IntentVCNet that unifies the temporal and spatial understanding knowledge inherent in LVLMs to bridge the spatio-temporal gap from both prompting and model perspectives.
Specifically, we first propose a prompt combination strategy designed to enable LLM to model the implicit relationship between prompts that characterize user intent and video sequences.
We then propose a parameter efficient box adapter that augments the object semantic information in the global visual context so that the visual token has a priori information about the user intent.
The final experiment proves that the combination of the two strategies can further enhance the LVLM's ability to model spatial details in video sequences, and facilitate the LVLMs to accurately generate controlled intent-oriented captions.
Our proposed method achieved state-of-the-art results in several open source LVLMs and was the runner-up in the IntentVC challenge. Our code is available on \url{https://github.com/thqiu0419/IntentVCNet}.
\end{abstract}



\maketitle

\section{Introduction}

\begin{figure}
   \centering
   \includegraphics[width=1.0\linewidth]{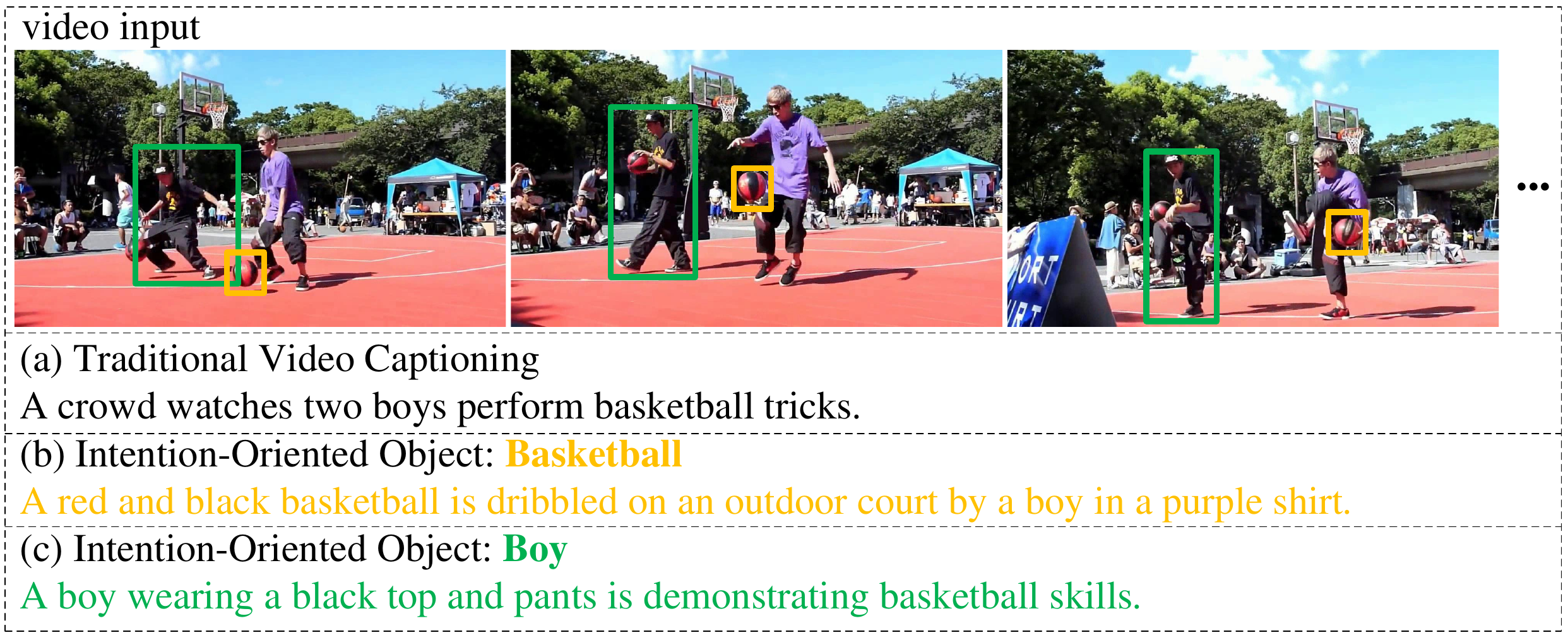}
   \caption{ Comparisons between the traditional video captioning task and the intention-oriented controllable video captioning task. (a) Traditional video captioning provides a general overview of the content but lack the specificity to address particular user needs. (b,c) Intention-oriented controllable video captioning emphasizes objects of user interest while accounting for the contextual information, resulting in more detailed and targeted captions.}
   \label{intro}
 \end{figure}

Video captioning, aiming at automatically generating a description of given videos, has attracted a lot of attention due to its potential to enhance visual understanding across both spatial and temporal dimensions. As shown in Fig. \ref{intro}, traditional video captioning prioritizes the accuracy and generality of description, focuses more on the overall understanding of the video, and is difficult to focus on the objects of interest to the user, which makes traditional video captioning perform poorly in personalized, highly accessible scenarios.
Therefore, introducing intention-oriented controllable caption generation is of significant value, which enables customized generation aligned with intention-oriented object and facilitates more personalized human-computer interaction experiences.

Intention-oriented controllable video captioning necessitates the tracking of the object of interest throughout dynamic video streams, posing challenges in comprehending both the regional object in each static frame and the corresponding temporal actions.
Recent studies have witnessed its great development which is primarily reflected in the aspects of large vision-language models (LVLM). LVLMs \cite{li2023blip,zhu2023minigpt,liu2023visual} expand the knowledge of large language models \cite{chiang2023vicuna,yang2025qwen3} (LLM) into visual domain, demonstrating the remarkable performance across various image-level tasks, including image captioning. Subsequent works delve into a more nuanced understanding towards spatial and temporal dimensions. In the spatial dimension, studies \cite{peng2023kosmos, chen2023shikra, zhang2025gpt4roi,you2023ferret,xie2024tune} integrate explicit positional information into LVLMs to enable regional tasks, such as visual grounding. They design various positional referencing methods to enhance fine-grained region comprehension, 
In the temporal dimension, \cite{zhang2025videollama, bai2025qwen2, wang2024internvideo2} employ video instruction tuning to adapt the model to video formats and to effectively model temporal relationships, exhibiting excellent performance on video captioning. Given the limited context length of LVLMs, they also explore to compress redundant visual tokens within frame sequences.

Although LVLMs demonstrate promising results on spatial understanding and video captioning respectively, there remains a spatio-temporal gap when tracking fine-grained objects across frame sequences. This limitation hinders the fine-grained controllability of LVLMs in intention-oriented controllable video captioning. This issue arises because current LVLMs acquire temporal modeling capabilities through pre-training on simplistic video-level instruction datasets, while they develop spatial understanding through pre-training on static images. There exists the spatio-temporal gap to bridge the static spatial understanding and dynamic temporal modeling. CAT-V \cite{tang2025caption} integrates the LVLMs with other experts in object recognition and temporal analysis to facilitate object-centric captioning. Nevertheless, CAT-V is a training-free framework, and consequently, its performance is constrained by the effectiveness of the various expert modules. Additionally, in CAT-V, the LVLM functions only as a basic captioner, leaving the spatio-temporal gap unaddressed. Therefore, current LVLMs still struggle to understand more fine-grained temporal changes pertaining to specific object.

To remedy the spatio-temporal gap, we propose the IntentVCNet, a spatio-temporal enhanced multi-modal collaborative framework. We substantially improve the fine-grained spatial understanding of LVLMs by advancing both prompt learning techniques and model architecture.
On one hand, instead of utilizing a single positional representation \cite{peng2023kosmos,chen2023shikra,ma2024groma}, we enhance the spatial modeling of fine-grained objects in LLM through the combination of prompts. On the other hand, we develop a global-local interaction module within the visual encoder to effectively extract region-enhanced visual features.
Additionally, we perform the parameter-efficient video instruction tuning to preserve the inherent vision-language knowledge and improve the LVLM's capacity to comprehend dynamic changes of intention-oriented object within videos.
Ultimately, we integrate the results from these models using a collaborative voting mechanism to improve overall performance.





Specifically, for the prompt combination, we fuse sequences of numerical coordinates in linguistic instruction and visual prompting in the videos, which enhances the fine-grained object localization from both visual and linguistic domains and acquires various heterogeneous models. The numerical coordinates of object is normalized in the instruction corresponding to each frame. For visual prompting, the intention-oriented object is highlighted by a red box in each frame. 
At the model level, we employ the robust InternVL3 \cite{zhu2025internvl3} and InternVideo2.5 \cite{wang2025internvideo2} as our foundational models. 
InternVL3 facilitates the processing of high-resolution videos, thereby ensuring the complete retention of visual information in each frame. In contrast, InternVideo2.5 implements efficient visual semantic compression to reduce redundant tokens, thereby enhancing its adaptability for longer video comprehension. 
To boost spatial interactions between intention-oriented objects and frame images, we propose a box adapter that incorporates global-local interaction modules. These modules facilitate the integration of object semantics into the global features of the frame. Finally, 
to achieve a synergistic result, we implement a collaborative voting process based on the textual similarity of descriptions generated by multiple heterogeneous models.
 
Our contributions can be summarized as follows: 

$\bullet$  We propose a prompt combination approach, which fuses the effective positional referring in both instruction and video data, improving the spatial modeling capacity of LLM to identify intention-oriented objects.

$\bullet$ We propose a parameter-efficient box adapter to boost spatial interaction between intention-oriented object and frame images, which acquires the region-enhanced visual features.


$\bullet$ We conduct extensive experiments on the IntentVC benchmark and achieve outstanding performance with 225.19\% CIDEr score on the test set, ranking 2nd in the IntentVC Challenge in conjunction with ACM MM'25.


\begin{figure*}
  \centering
  \includegraphics[width=0.8\textwidth]{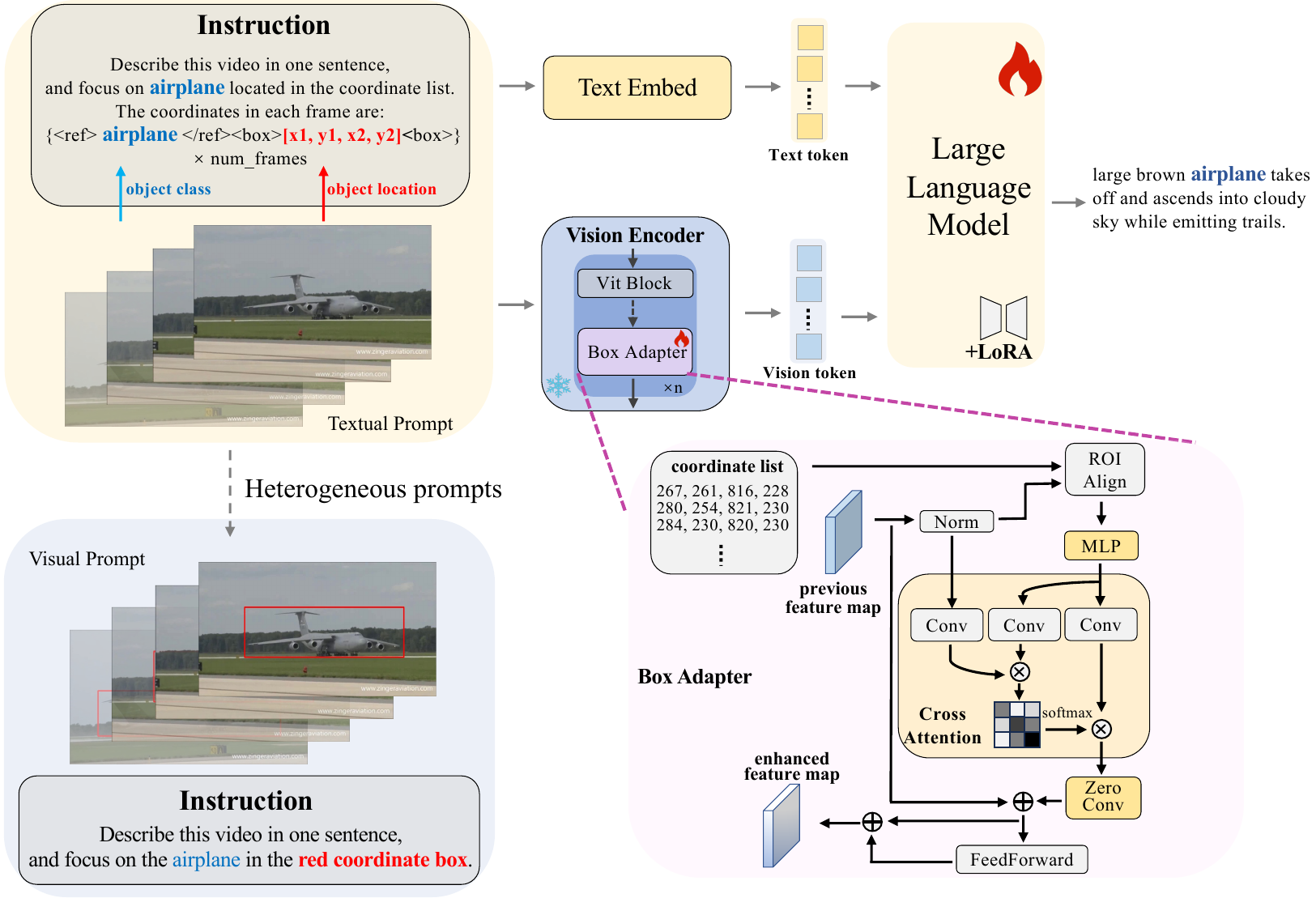}
  \caption{An overview of our framework for intention-oriented video captioning. 1) We first design a prompt combination, which incorporates the linguistic coordinates and visual prompting. 2) In vision encoder, we insert the box adapter to enhance the regional visual information through a global-local interaction. 3) Finally, original visual encoder is frozen, and only the lightweight box adapter is optimized. Additionally, the LLM is trained using LoRA \cite{hu2022lora}. }
  \label{fig_framework}
\end{figure*}

\section{Related Work}

\subsection{Video Captioning}  
Video captioning (VC) has witnessed remarkable advancements, evolving from early sophisticated neural architectures to large vision-language models. These works leverage encoder-decoder frameworks, where visual encoders (CNNs/ViTs) extract visual features and textual decoders (RNNs/Transformers) generate captions. Early efforts employ the attention mechanisms \cite{chen2017video,li2017mam,lin2022swinbert}, graph networks \cite{zhang2019object,xiao2020video} and reinforcement learning \cite{pasunuru2017reinforced,wang2018video,lyu2023macaw}. With the rise of pre-training techniques, subsequent works \cite{sun2019videobert,xue2022clip,tong2022videomae,ye2023hitea} follow the "pretraining-finetuning" paradigm. The pretrained model can be fine-tuned to accommodate various downstream tasks, including video captioning. Recently, LVLMs have advanced rapidly. Many works also explore the use of LVLMs in video understanding, which obtain a versatile model capable of performing various tasks. They persistently optimize the spatio-temporal  interaction \cite{chen2023videollm,cheng2024videollama,lin2023video,lyu2023macaw} and training strategy \cite{zhang2025videollama,shu2023audio,xu2024slowfast} to enhance the temporal modeling capabilities of fundamental LVLMs. InternVL \cite{zhu2025internvl3,wang2025internvideo2} and QwenVL \cite{bai2025qwen2} represent cutting-edge models in the domain of video understanding, particularly in video captioning.

With the increasing ability of human-computer interaction systems, the demand for captions that are not only descriptive but also tailored to specific user intentions. This evolution has given rise to controllable video captioning. 
Controllable signals can be primarily classified into two categories: structural control and content control. The former regulates the grammatical structure \cite{wang2019controllable,sun2024syntax} of generated sentences, whereas the latter constrains the content, encompassing objects \cite{zhu2020ovc,yao2024edit}, relations \cite{cao2025skeleton}, and emotional aspects \cite{song2023contextual,song2023emotion,ye2024dual,song2024emotional}. For object-oriented control, OVC-Net \cite{zhu2020ovc} proposes a temporal graph to emphasize specific objects.  Elysium \cite{wang2024elysium} and GroundingGPT \cite{li2024groundinggpt} construct the object-level instruction datasets and achieve promising performance on grounding task. 
However, due to the spatio-temporal gap from scarcity of training data and insufficient model adaptation, it is still not possible to fully leverage them for object-oriented controllable video captioning.

\subsection{Spatial Understanding in LVLMs}

To enhance the spatial understanding of the visual world through LVLMs, various positional representations have been proposed in existing literature. Kosmos \cite{peng2023kosmos} was the first to introduce a unified positional representation method by employing specialized location tokens to signify regions. Shikra \cite{chen2023shikra} further streamlined earlier approaches by directly utilizing numeric coordinates for representation. GPT4RoI \cite{zhang2025gpt4roi} increases the importance of object-level region features in interactions from a feature perspective. Ferret \cite{you2023ferret} consolidates prior representations and introduces a hybrid spatial representation approach that incorporates triples, consisting of region names, numeric coordinates, and region features to define a region. A region is defined by a four-dimensional coordinate system, represented by the upper left and lower right points. The method described previously incorporates positional representations into linguistic instructions. However, within the current paradigm of LVLMs, this approach consumes a substantial portion of the available context length, which may result in window overflow and a decline in model performance. Furthermore, \cite{xie2024tune} has shown that the visual encoder in LVLMs is particularly sensitive to visual markers. Consequently, these special markers \cite{yang2023set} can also serve as visual prompts, which does not add additional positional token length to the context. 

\section{Method} \label{section_3}
Our proposed model is illustrated in Fig.~\ref{fig_framework}. 
Technically, from the prompting perspective, we begin by designing a prompt combination approach, where the numerical coordinates within linguistic instructions and visual prompts within the videos are combined, thereby enhancing fine-grained object localization of LLM and acquiring various heterogeneous models. In terms of visual prompting, the target object of interest is prominently highlighted with a red box in each frame. From the model perspective, to enhance spatial interactions between the target objects and the frame images, we propose a box adapter that incorporates cross-attention modules. These modules enable the integration of object semantics into the global features of the frame. Finally, we introduce a multi-model collaborative strategy designed to integrate various models for videos of differing lengths.

\subsection{Pre-trained Large Vision-Language Models} \label{section_3_1}
Large vision-language models are developed based on LLMs and are continuously pre-trained using extensive video instruction data, which exhibits remarkable capabilities of video understanding and instruction following. In this paper, we utilize InternVL3 \cite{zhu2025internvl3} and InternVideo2.5 \cite{wang2025internvideo2} to analyze videos of varying lengths.

\textbf{InternVL3} comprises three modules: a visual encoder, a multimodal connector, and a LLM. The input video frames are initially partitioned into image tiles. 
Subsequently, a fixed-resolution visual encoder is employed to extract their visual features, thereby supporting dynamic high-resolution to maximize the retention of visual information. The multimodal connector consists of a MLP layer and pixel unshuffle operation, which projects the visual content into the representation space of LLMs and streamline the visual embeddings. 
These visual features are then positioned within the designated slots of the embedded linguistic instructions, collectively forming the context embeddings of LLM. 


\textbf{InternVideo2.5.} Building on the InternVL foundational model, InternVideo2.5 advances through post-training for long video data. InternVideo2.5 additionally implements hierarchical vision token compression based on semantic similarity of visual features, enabling the model to incorporate more video frames within a constrained context length, thus achieving long-range video modeling. 
Additionally, in terms of training strategy, InternVideo2.5 employs direct preference optimization to enhance dense visual tasks. 

\subsection{Prompt Combination} \label{section_3_2} 
Previous studies \cite{peng2023kosmos,chen2023shikra,zhang2025gpt4roi,you2023ferret} have employed various positional referencing methods within the instructions to facilitate the model's understanding of specific regions. 
In this paper, we propose a prompt combination approach within the user instruction and the visual input respectively. Through designing combined positional prompts, the LLM attains fine-grained spatial modeling capacity, allowing for its extension to various heterogeneous models. Specifically, the prompt combination contains the numeric coordinates in instruction and visual prompting.

\ding{172} \textbf{The numeric coordinates in instruction.} LVLMs offer controllability through user instructions, which incorporate users' intentions, making them essential for intention-oriented video captioning. In this paper, our controllable element is a specific object, while objects in video data are constantly moving and changing. Therefore, simple textual instructions cannot adequately serve as a reference for the intention-oriented object. We extend the approach of numeric coordinates from spatial understanding of static image to dynamic video.
Specifically, we map the coordinates of the object regions of interest in each frame to their respective frames in textual format. The coordinates are represented as four-dimensional vectors, specifically indicating the horizontal and vertical coordinates of the upper-left and lower-right locations, denoted as $[x_1, y_1, x_2, y_2]$. To standardize various sizes, these values are normalized to a range of 0 to 1000, and the resulting user instructions. 

\ding{173} \textbf{Visual Prompting.} \cite{xie2024tune,yang2023set} have demonstrated that the visual encoder of LVLMs is particularly sensitive to specific salient visual markers. Consequently, subsequent studies have sought to highlight intended reference areas by incorporating visual markers into images. These markers serve as visual prompts and can also be effectively extended to video data. We visualize the coordinates of intention-oriented object onto the corresponding video frame. As illustrated in the Fig. \ref{fig_framework}, the red rectangular areas denote our visualization results for these coordinates. It is important to note that, in comparison to the original coordinate size, we have slightly enlarged the range of the bounding box to minimize excessive obstruction of the target objects within the red box.


\subsection{Box Adapter} \label{section_3_3}

Current LVLMs aim at enhancing spatial understanding demonstrate inadequate interaction with specific regions. They have acquired extensive multimodal knowledge through pre-training, which is embedded within their parameters. Consequently, directly altering the model structure to improve fine-grained regional interaction may jeopardize the intrinsic knowledge. 
Previous works induce the Parameter-Efficient Fine-Tuning (PEFT) methods such as prefix tuning \cite{li2021prefix}, adapter tuning \cite{houlsby2019parameter} and LoRA \cite{hu2022lora}, which freezes the original LVLMs and inserts a limited number of trainable new parameters, thereby facilitating model fine-tuning while preserving the knowledge acquired from the pre-trained model. Inspired from these PEFT methods, we propose the box adapter, which is integrated into the original LVLMs to enhance the deeper interaction with the intention-oriented object.

Specifically, as shown in Fig. \ref{fig_framework}, given a visual feature map $V_f = \{v_{fi} \in \mathbb{R}^{d \times h \times w}\}_{i=1}^{N_v}$ of the $i$-th frame, a box adapter firstly extracts the region features of the intention-oriented object through Region-of-Interest (RoI) alignment, which can be presented as:
\begin{equation}
R=\operatorname{RoI\_Align}\left(\operatorname{LN}\left(V_f\right), bbox\right), \\
\end{equation}
where the $bbox$ is the numerical coordinates of intention-oriented object and $R \in \mathbb{R}^{N_v \times d \times h' \times w'}$ represents its region features. Then, we perform the global-local interaction through a cross-attention module. The complete visual feature map $V_f$ functions as the query embeddings while the region features serve as the key-value embeddings. This design injects the regional visual information into overall visual features, thereby establishing spatial associations between global and local visual elements. Formally, given region features $R$ and visual feature map $V_f$, it is formulated as:
\begin{equation}
\begin{aligned}
& \tilde{V}_f=V_f+\mathbb{Z}\left(\operatorname{MHA}\left(Conv_Q(V_f), Conv_K(R),Conv_V(R)\right)\right), \\
&V_{fr}=\tilde{V}_f+\operatorname{FFN}\left(\operatorname{LN}\left(\tilde V_{f}\right)\right),
\end{aligned}
\end{equation}
where $\operatorname{MHA}$, $\operatorname{LN}$, and $\operatorname{FFN}$ denote multi-head attention, layer normalization, and feed-forward networks, respectively. The $Conv_Q$, $Conv_K$, $Conv_V$ are the 1*1 convolutions and they are responsible for the projection to get query, key and value. 
$\mathbb{Z}$ denotes zero conv, inspired by~\cite{zhang2023adding}, we introduce zero conv with weight and bias initialized to 0 to prevent the instability brought by the preliminary training.
The final $V_{fr} \in \mathbb{R}^{N_v \times d \times h \times w}$ is the region-enhanced visual feature map. 

To promote the deep interaction between global and local visual information, we incorporate the box adapter into the visual encoder of the LVLMs. Deeper visual features inherently contain more high-level semantic information, so we insert the box adapter into several deeper layers of the visual encoder. The InternVL series models utilize the Vision Transformer \cite{dosovitskiy2020image} (ViT) as their visual encoder. Consequently, we position the box adapter behind the ViT layers, progressively enhancing the local object information of the visual features. 
The global-local deep fusion result produces as:
\begin{equation} \label{equ_fusion}
\begin{aligned}
V_f^{(\tilde l)} & =\operatorname{ViT\_Layer}\left(V_f^{(l)}\right), \\
V_f^{(l+1)} & = V_{fr}^{(l)} =\operatorname{Box\_Adapter}\left(V_f^{(\tilde l)}\right).
\end{aligned}
\end{equation}
The visual feature map denoted as ${V}_f^{(l)} \in \mathbb{R}^{N_v \times d \times h \times w}$ is fed into the $ l $-th layer of the ViT. Consequently, the final region-enhanced visual features effectively mitigate the spatio-temporal gap from model perspective.

\subsection{Multi-Model Ensemble Collaboration} \label{section_3_4}
After the video instruction tuning, we obtain the heterogeneous models from the foundation models InternVL3 and InternVideo2.5. Inspired by \cite{li2024place}, we develop a collaborative voting mechanism to  integrate the descriptive results from multiple models. Specifically, we compute the text similarity among the descriptions generated by multiple models. The similarity score can be obtained through various methods, including cosine similarity of sentence-level text embeddings and matching scores at the word or character level. We select the sentence with the highest average similarity score as the final description. A high average similarity indicates that multiple models have reached a consensus, suggesting that this sentence most accurately reflects the input video.

%
\section{Experimental Results}
\subsection{Dataset}
We use the official dataset provided by IntentVC Challenge\cite{intentvc2025}, which is labeled based on the LaSoT dataset\cite{fan2019lasot}. The dataset has a total of 70 different categories as specific user intents, and each category contains 20 videos of different objects. More specifically, the FPS of each video is set to 1 and each video frame has a unique visual grounding annotation for its corresponding object in the standard COCO format like [x,y,w,h]. When the object is moved out of the scene, its corresponding grounding box is set to [0,0,0,0,0]. The training set, the public test set, and the private test set are divided in the order of 14:3:3, where each video in the training set has five fine manually labeled captions.

\subsection{Implementation Details.}\label{imp}
All of our experiments were realized in Pytorch 2.1.1 and CUDA 12.1 environments using 4 NVIDIA H100 80G GPUs. During training, we freeze the visual extractor and then train the LLM with the lora strategy with rank=128. For each ablation experiment, we use AdamW optimizer ($\beta_1=0.9$, $\beta_2=0.999$ and $weight\_decay=0.05$) with a batch size of 16. The initial value of the learning rate was $2 \times 10^{-5}$ and is updated by a cosine annealing schedule. The training image size is force set to $448 \times 448$ pixels. For data augmentation, we only use a random sampling strategy in the time dimension, where 32-48 frames are randomly sampled during training and 48 frames are fixedly used during inference.

\subsection{Evaluation Metrics.}
Follow IntentVC challenge, we will use the four most commonly employed metrics for evaluating video captioning: BLEU@4~\cite{papineni2002bleu}, METEOR~\cite{banerjee2005meteor}, CIDEr~\cite{vedantam2015cider}, and ROUGE-L~\cite{rouge2004package}. 

\subsection{Comparison with State-of-the-Art Methods}
\begin{table}[]
\setlength{\tabcolsep}{3pt}
\caption{Comparison with state-of-the-art methods.}
\label{table_pretain}
\begin{tabular}{l||cccc}
\toprule
Method   & CIDEr & METEOR & BLEU@4 & ROUGE-L \\ 
\midrule
VAST~\cite{chen2023vast} & 139.41 & 47.67 & 23.44 & 47.12 \\
Qwen2.5-VL~\cite{bai2025qwen2} & 165.61 & 53.31 & 30.30 & 51.45 \\
InternVideo2.5~\cite{wang2025internvideo2} & 187.48 & 56.39 & 36.41 & 54.80 \\
InternVL3~\cite{zhu2025internvl3} & 186.71 & 54.79 & 38.79  & 56.09 \\
\midrule
IntentVCNet(Ours) & \textbf{225.19} & \textbf{62.36} & \textbf{45.09} & \textbf{60.07} \\
\bottomrule
\end{tabular}
\label{tab:qr}
\end{table}
To validate the effectiveness, we compared the proposed method with some advanced LVLM methods on the IntentVC public test set.
The qualitative comparison results are shown in Table~\ref{tab:qr}. We selected four state-of-the-art open-source LVLM including VAST~\cite{chen2023vast}, Qwen2.5-VL~\cite{bai2025qwen2}, InternVideo2.5~\cite{wang2025internvideo2}, InternVL3~\cite{zhu2025internvl3} and then fine-tuned them on the IntentVC dataset for fair comparison. As shown in Table~\ref{tab:qr}, our proposed method achieves the best results on CIDEr, METEOR, BLEU@4, ROUGE-L, which proves the effectiveness of our proposed method and strategy. Even though InternVideo is a generative large model which focuses on the video domain, the method we proposed still outperforms it by 37.71 on the CIDEr, and all the other metrics are also considerably improved.

\subsection{Analysis and Discussion}
In this section, we provide an in-depth analysis and demonstrate the effectiveness of each of our proposed components. Table~\ref{tab:ablation} shows the ablation experiments for each component using InternVL3 as the baseline. For the sake of brevity and comprehensibility, we show only the metrics for the public test set, with roughly comparable trends for the private test set.

\begin{table}[htbp]
\centering
\caption{Ablation experiments. TP,VP,BA stand for textual prompts, visual prompts, and box adapter, respectively. All experiments use beam search strategy of length 5 for inference, and the rest of experimental setup is the same as Sec.~\ref{imp}}
   \begin{tabular}{ccc||cccc}
      \toprule
      TP & VP & BA & BLEU@4 & METEOR & CIDEr & ROUGE-L \\
      \midrule
                 &            &            & 40.56 & 56.97 & 196.2  & 58.01 \\
      \checkmark &            &            & 43.45 & 58.54 & 211.45 & 59.02 \\
                 & \checkmark &            & 43.22 & 58.88  & 210.76 & 58.89 \\
      \checkmark & \checkmark &            & 42.17 & \underline{59.84} & 214.45 & 58.43 \\
      \midrule
                 &            & \checkmark & 42.19 & 57.73  & 204.71 & 58.02 \\
      \checkmark &            & \checkmark & \textbf{44.98} & \textbf{60.67} & \textbf{223.01} & \textbf{60.7} \\

      \checkmark & \checkmark & \checkmark & \underline{43.72} & 59.29 & \underline{217.17} & \underline{59.08} \\
      \bottomrule
   \end{tabular}
   \label{tab:ablation}
\end{table}

\begin{table}[htbp]
\centering
\caption{Experiments on box adapter position. "embed layer" denotes the embedding part after vision model, the rest denotes the incorporation of the box adapter in the last n layers. }
   \begin{tabular}{c||cccc}
      \toprule
      Settings & BLEU@4 & METEOR & CIDEr & ROUGE-L \\
      \midrule
      baseline & 43.45 & 58.54 & 211.45 & 59.02 \\
      \midrule
      +embed layer    & 43.79 & \underline{59.64} & 217.74 & \underline{59.96} \\
      +last 3 layers & \underline{43.79} & 59.31 & \underline{219.54} & 59.62 \\
      +last 5 layers  & \textbf{44.98} & \textbf{60.67} & \textbf{223.01} & \textbf{60.7} \\
      +last 8 layers  & 42.14 & 58.32 & 206.94 & 58.2 \\
      +last 9 layers & 42.22 & 57.87 & 205.93 & 58.48 \\
      
      \bottomrule
   \end{tabular}

   \label{tab:ba}
\end{table}


\begin{table}[htbp]
\centering
\caption{Integration experiment. We simply let InternVL process shorter videos, InternVideo process longer videos, and finally concatenate the results.}
\setlength{\tabcolsep}{3pt}
   \begin{tabular}{c||cccc}
      \toprule
      Settings & BLEU@4 & METEOR & CIDEr & ROUGE-L \\
      \midrule
      InternVL3~\cite{zhu2025internvl3} & 43.45 & 58.54 & 211.45 & 59.02\\
      InternVideo2.5~\cite{wang2025internvideo2} & 42.77 & \textbf{61.37} & 215.62 & 59.00 \\
      fusion & \textbf{44.28} & 61.01 & \textbf{221.0} & \textbf{59.96} \\
      \bottomrule
   \end{tabular}
   \label{tab:fuse}
\end{table}



\noindent\textbf{Prompt combinations.}
As shown in Table~\ref{tab:ablation}, the two different modalities of prompts can provide considerable performance improvement to the baseline, indicating that reasonable prompts can significantly improve the model's attention to the intention of the user, and can effectively guide the LLMs to generate text that matches the intention.
However, combining visual and textual prompts did not result in the expected large improvement, and the model only showed a small improvement in the CIDEr (211.45 → 214.45). We suggest that this is due to the fact that either prompt was sufficient to improve the model's ability to attend to the target, whereas using them together leads to redundancy, which in turn triggers overfitting.
Thus we split the visual and textual prompts as heterogeneous models to participate in the final ensemble, rather than using them both in a single model.

\noindent\textbf{Box adapter.}
After the introduction of box adapter, the model's ability to understand the intention of the user is further improved.
Specifically, the performance of CIDEr improves from 211.45 to 223.01 compared to the model using textual prompts, which demonstrates the effectiveness of box adapter for controlled video captioning. In addition, since box adapter can be dynamically integrated into the vision extractor, Table~\ref{tab:ba} shows a comparison of the effectiveness of adding box adapter at different levels. From the experimental results, incorporating box adapter in too many layers will not only make the network bulky, but the accuracy will also be affected due to overfitting. Weighing the pros and cons, we choose to incorporate the box adapter in the last five layers of the vision model, and the CIDEr can reach the highest 223.01.


\noindent\textbf{Fusion necessity.}
We chose two LVLMs mainstream in the video domain as baseline, where InternVL is suitable for processing short videos and InternVideo is able to process longer videos due to the use of token compression strategy. In order to verify the necessity of fusion, we manually truncate each video, and videos smaller than 74 frames are processed by utilizing InternVL and vice versa by utilizing InternVideo, and the experimental results are shown in Table~\ref{tab:fuse}. From the results, it is clear that simple fusion according to the comfort zone of the model can also be effective in improving the accuracy of the model, which drives us to use the voting strategy to fuse more models in the end.

\section{Conclusion}
In this paper, we propose IntentVCNet, a novel framework for intention-oriented controllable video captioning that addresses the fundamental spatio-temporal gap in existing Large Vision-Language Models. Our approach tackles the core challenge of generating user-controllable, intention-oriented captions by bridging static spatial understanding with dynamic temporal modeling. First, we introduce a prompt combination strategy that fuses numerical coordinates in linguistic instructions with visual prompting in video data, enabling fine-grained object localization across both visual and linguistic domains. Second, we develop a parameter-efficient box adapter that enhances spatial interactions between intention-oriented objects and frame images through global-local feature fusion. Our method can generate targeted, intention-oriented captions that focus on specific objects while maintaining contextual coherence, representing a significant advancement in controllable video understanding. Future work will explore extending our approach to multi-object intention control and investigating more sophisticated temporal modeling strategies for long-form video content.

\section{ACKNOWLEDGMENTS}
This work was supported in part by the China Postdoctoral Science Foundation (2025M771515 ) and Anhui Postdoctoral Scientific Research Program Foundation (2025C1166). The computational work in this paper was supported by the technical assistance of the Network Information Center and the Smart Campus Project at the University of Science and Technology of China. We gratefully acknowledge their support.

\bibliographystyle{ACM-Reference-Format}
\bibliography{main}

\end{document}